# An Exploratory Study of Masked Face Recognition with Machine Learning Algorithms

Megh Pudyel
Department of Computer Science
Winston-Salem State University
Winston-Salem, NC USA
mpoudyel119@rams.wssu.edu

Mustafa Atay
Department of Computer Science
Winston-Salem State University
Winston-Salem, NC USA
ataymu@wssu.edu

*Abstract*— Automated face recognition is a widely adopted machine learning technology for contactless identification of people in various processes such as automated border control, secure login to electronic devices, community surveillance, tracking school attendance, workplace clock in and clock out. Using face masks have become crucial in our daily life with the recent world-wide COVID-19 pandemic. The use of face masks causes the performance of conventional face recognition technologies to degrade considerably. The effect of mask-wearing in face recognition is yet an understudied issue. In this paper, we address this issue by evaluating the performance of a number of face recognition models which are tested by identifying masked and unmasked face images. We use six conventional machine learning algorithms, which are SVC, KNN, LDA, DT, LR and NB, to find out the ones which perform best, besides the ones which poorly perform, in the presence of masked face images. Local Binary Pattern (LBP) is utilized as the feature extraction operator. We generated and used synthesized masked face images. We prepared unmasked, masked, and half-masked training datasets and evaluated the face recognition performance against both masked and unmasked images to present a broad view of this crucial problem. We believe that our study is unique in elaborating the mask-aware facial recognition with almost all possible scenarios including half_masked-to-masked and half_masked-to-unmasked besides evaluating a larger number of conventional machine learning algorithms compared the other studies in the literature.

I. INTRODUCTION

Traditional facial recognition systems have performed fairly well with unmasked faces until the COVID-19 pandemic dominated the whole world. The governments made wearing masks or face coverings as mandatory means of protection to prevent the spread of contagious COVID-19 virus in most of the countries. Only trained to work with unmasked faces, these facial recognition systems then started to degrade in performance for identifying masked face images [5]. The occluded faces have been identified to affect the face recognition solutions while developing occlusion invariant facial recognition solutions have become growing research challenges [13][14]. These solutions have been the outcomes of the models trained with either unmasked or masked faces.

Various studies have been done for occluded facial recognition solutions, but none of them have made a comparative study by using masked, unmasked and half-masked training datasets and a large number of conventional Machine Learning (ML) models to better explore strengths and weakness of those models, to the best of our knowledge. We train and test each model with all masked, unmasked, and half-masked faces. It is essential to make a comparative study to find out the high performing models in each one of these cases of unmasked, masked, and half-masked trained models, and report the ones which are high performers besides the poor ones.

In this study, 6 conventional ML models are first trained with unmasked, masked, and half-masked face images and tested with unmasked and masked face images. We explore and report the ML models with best performance in each experimental setting when trained with unmasked, masked, and half-masked datasets. The 6 conventional ML models that we experimented are Support Vector Classifier (SVC), Linear Discriminant Analysis (LDA), K-Nearest Neighbors (KNN), Decision Trees (DT), Logistic Regression (LR) and Naïve Bayes (NB). We also tracked and reported miss rates for masked and unmasked images besides reporting performance metrics such as accuracy and F1 score.

II. RELATED WORK

A number of algorithms have been developed for non-masked face recognition which are widely used and show good performance. Nonetheless, not so many contributions have been made in the field of masked face recognition.

Dharanesh et al. [4] proposed a solution for recognizing face in the presence of mask by using dynamic ensemble of deep learning models. They use face or ocular regions for recognition depending upon masked or unmasked faces. They propose switching to ocular region processing in run-time for testing face images in the presence of mask. Their experimental results suggested that the proposed solution in the presence of facial mask obtains comparable performance to the conventional face recognition system in the absence of the mask.

Damer et al. [5] performed an exploratory analysis of face recognition system considering the effect of wearing mask in the recognition performance by studying two non-commercial models, namely, ArcFace and SphereFace, and one commercial off-the-shelf (COTS) model, namely, MegaMatcher 11.2 SDK. The evaluation scores of genuine and imposter comparisons are calculated in all the three systems considered. The effect of mask is found apparent in all the three systems. The effect is found most significant on the genuine score's distribution rather than the imposter scores distribution. Therefore, they state that the current face recognition solutions are not promising enough to match masked faces with unmasked faces. They need to be re-evaluated for proper performance when considering masked faces.

In another experiment, Damer et al. [10] compared the performance of automated Face Recognition (FR) against FR by human experts. Their work hints the possibility of enhancing the masked face verification performance of human experts through explicit training. They also provide important clues for the development of FR solutions that are robust to masked faces like training FR models that can process both masked and unmasked faces or reducing the effect of the mask on the face embedding by learning to transfer it into an embedding that behaves similarly to that of an unmasked face.

Montero et al. [6] proposed an approach taking the ArcFace model designed by Deng et al. with several modifications for the backbone and the loss function, converting it to Multi-Task ArcFace model. Their experiments showed that the proposed approach highly boosts the original model's accuracy when dealing with masked faces. They preserve almost the same accuracy on the original non-masked datasets in mask-usage classification.

Anwar et al. [7] address a methodology to use the current facial datasets by augmenting it with an open-source tool called MaskTheFace. The MaskedTheFace tool enables masked faces to be recognized with low false-positive rates and high accuracy without requiring the user dataset to be recreated by taking new pictures for authentication. They report an increase in the true positive rate for the FaceNet system. They also test the accuracy of re-trained system on a custom real-world dataset MFR2 and report similar accuracy.

Ejaz et al. [8] used Principal Component Analysis (PCA), a successful statistical and widely used tool applied in non-masked face recognition and applied in the masked face recognition problem. They performed a comparative study for a better understanding using ORL face database. They concluded that PCA gives poor recognition rate for masked face images compared to non-masked faces.

While we evaluate masked face recognition with 6 conventional ML algorithms and 3 different types of training datasets in this paper, none of the presented work in the literature evaluated as many algorithms and scenarios as studied in this paper, to our knowledge.

III. PRELIMINARIES

We introduce 6 conventional ML algorithms that we use in our experiments as well as the ORL image database along with MaskTheFace software for synthesizing masked face images in the following.

A. Machine Learning Algorithms

Classification algorithms are used to classify objects of various types. They help to classify objects into similar or dissimilar groups. These algorithms also play an integral role in facial recognition. They help to categorize the images and determine their relationship to each other. Our exploratory study uses a total of 6 different conventional machine learning classification algorithms for experimentation [15, 17]. They are Support Vector Classifier (SVC), Linear Discriminant Analysis (LDA), K-Nearest Neighbors (KNN), Decision Trees (DT), Logistic Regression (LR), and Naïve Bayes(NB).

- *Support Vector Classifier (SVC)* – Support Vector Classifier supports binary classification problems and can also be extended to handle multi-class problems. SVC maintains high generalization as it maps its inputs non-linearly to high-dimensional feature spaces and constructs linear decision surfaces.

- *Linear Discriminant Analysis (LDA)* – Linear Discriminant Analysis (LDA), and Principal Component Analysis (PCA) both are well-known classification techniques. As its name suggests, LDA is a linear classifier. LDA is very useful algorithm for dimensionality reduction. It is commonly used to extract features in pattern classification problems.

- *K-Nearest Neighbors (KNN)* – K-nearest neighbors (KNN) is used for solving mainly data mining and image classification problems. KNN is both classifier and regressor, but we use it in this paper as a classifier.

- *Decision Trees (DT)* – Decision Trees represent flowchart-like structure. Decision Trees are not like Support Vector Classifiers and neural networks as they do not make statistical assumptions concerning the inputs or scale of the data.

- *Logistic Regression (LR)* – Logistic Regression helps to model the probability of a specific class or existing classes. Despite its name, it is a classifier rather than being a regressor. It is a simple and very efficient method that we use for binary and linear classification problems. It is the most used ML model in the industry.

- Naïve Bayes (NB): Naïve Bayes is good for binary and multiclass classification problems. It is supposed to perform well in categorical input compared to numerical variables.

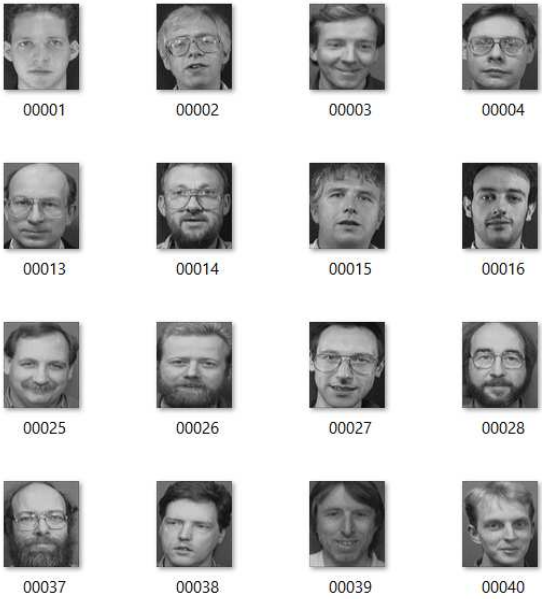

Fig.1. Sample images from ORL database

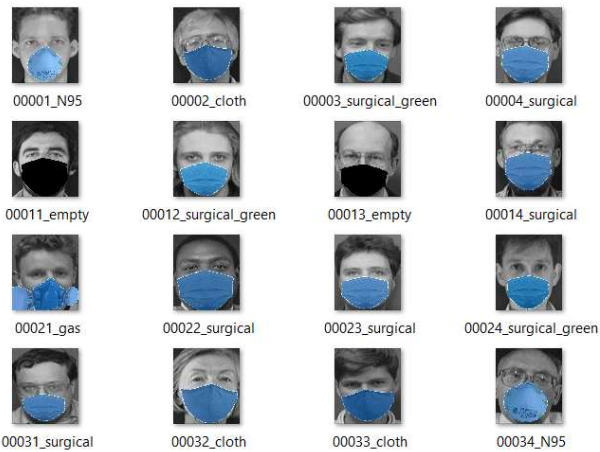

Fig.2. Sample masked ORL face images

### B. Database

We use the ORL (Our Database of Faces) database in our study [8] which is also used by Anwar et al. [7]. ORL database has 41 subjects and 10 images per subject with a total of 410 unmasked facial images. Sample images from ORL database is shown in Figure 1.

We use an open-source software, MaskTheFace to augment faces from the ORL database with masks. Sample masked ORL images which are created using MaskTheFace software are given in Figure 2. Table 1 shows the description of training and testing datasets which are extracted from ORL database and used in our experiments.

TABLE 1. TRAINING AND TESTING DATASETS

| DATASETS | Description | # of Subjects | # of Images/Subject | Total # of Images |
|---|---|---|---|---|
| Training_UM | All unmasked training images | 41 | 9 | 369 |
| Training_HM | Half unmasked & half masked training images | 41 | 8 | 328 |
| Training_M | All masked training images | 41 | 9 | 369 |
| Testing_UM | All unmasked testing images | 41 | 1 | 41 |
| Testing_M | All masked testing images | 41 | 1 | 41 |

### C. Synthesizing Masked Face Images

MaskedTheFace is a computer-vision based software which is used to synthesize masked face images. It uses a dlib based face-landmark detector to recognize face tilt. It has six mask templates to use from. In this paper, we use random selection of masks to synthesize our masked images from ORL database [16].

## IV. METHODOLOGY

We use 41 distinct subjects with 10 images each from the ORL database in our experiments. We performed six experiments using 6 conventional ML models. We utilized Local Binary Pattern (LBP) algorithm for feature extraction in all the experiments. We set up radius and neighborhood size to 8 and 24 respectively. We conducted the following experiments:

*Experiment 1:* We performed this experiment by training 6 ML models using 9 unmasked images of each 41 subjects in ORL database. Then we tested each of the ML models with 41

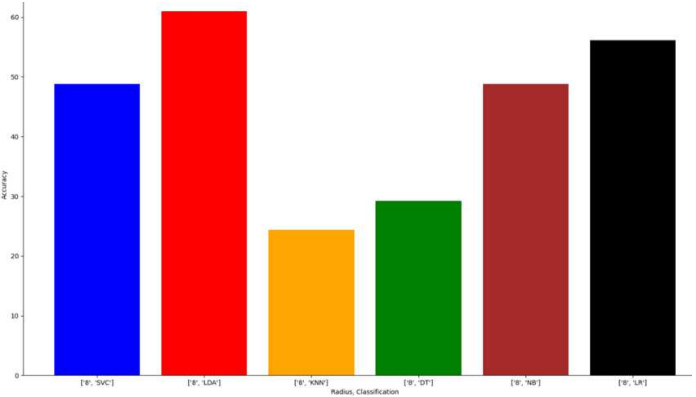

Fig. 3. Training_UM vs. Testing_M

unmasked images, 1 for each one of 41 individuals.

*Experiment 2:* We performed this experiment by training 6 ML models using 9 unmasked images of each 41 subjects. Then, we tested each of the models with 41 masked images synthesized by using the MaskTheFace software, 1 for each one of 41 individuals. Figure 3 shows the accuracy level of 6 ML models in this experiment.

*Experiment 3:* In this experiment, 6 ML models are trained with 9 masked images of each 41 individuals, and then tested each of the models with 41 unmasked images, 1 for each one of 41 individuals.

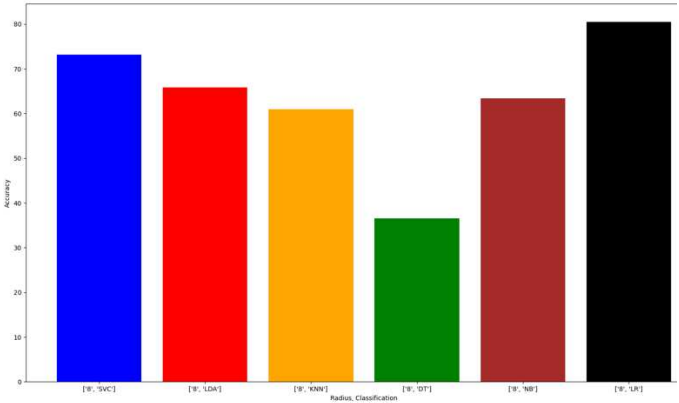

Fig.4. Training_M vs. Testing_M

TABLE 2. THE EXPERIMENTAL RESULTS SHOWING ACCURACY OF 6 ML MODELS IN ALL SIX EXPERIMENTS.

| Accuracy Table | Experiments | | | | | | AVERAGES |
|---|---|---|---|---|---|---|---|
| ML Algorithm | UM/UM | UM/M | HM/UM | HM/M | M/UM | M/M | |
| SVC | 85% | 49% | 80% | 71% | 59% | 73% | 70% |
| LDA | 90% | 61% | 76% | 76% | 63% | 66% | 72% |
| KNN | 78% | 24% | 59% | 37% | 17% | 61% | 46% |
| DT | 51% | 29% | 39% | 39% | 27% | 37% | 37% |
| LR | 98% | 56% | 80% | 80% | 71% | 80% | 78% |
| NB | 83% | 49% | 73% | 66% | 54% | 63% | 65% |
| AVERAGES | 81% | 45% | 68% | 61% | 48% | 63% | 61% |

*Experiment 4:* In this experiment, 6 ML models are trained with 9 masked images of each 41 individuals, and then tested each of the models with 41 masked images, 1 for each one of 41 individuals. Figure 4 shows the accuracy level of 6 ML models in this experiment.

TABLE 3. THE EXPERIMENTAL RESULTS SHOWING F1 SCORES OF 6 ML MODELS IN ALL SIX EXPERIMENTS.

| F1-Score Table | Experiments | | | | | | AVERAGES |
|---|---|---|---|---|---|---|---|
| ML Algorithm | UM/UM | UM/M | HM/UM | HM/M | M/UM | M/M | |
| SVC | 92.11% | 66% | 89% | 83% | 74% | 85% | 81% |
| LDA | 95% | 76% | 86% | 86% | 78% | 79% | 83% |
| KNN | 88% | 39% | 74% | 54% | 29% | 76% | 60% |
| DT | 68% | 45% | 56% | 56% | 42% | 54% | 54% |
| LR | 99% | 72% | 89% | 89% | 83% | 89% | 87% |
| NB | 91% | 66% | 85% | 79% | 70% | 78% | 78% |
| AVERAGES | 89% | 61% | 80% | 75% | 63% | 77% | |

*Experiment 5:* Out of 10 images of each 41 individual, 1 image is set aside for testing while from remaining 9 images, four images are masked with MaskTheFace software and combined with four unmasked images to make a total of 8 images. Thus, we make it half-masked images dataset. Then, this dataset containing 4 masked and 4 unmasked images of each 41 subjects is used to train 6 ML models. Then, we tested each of the models with 41 unmasked images, one for each one of 41 individuals.

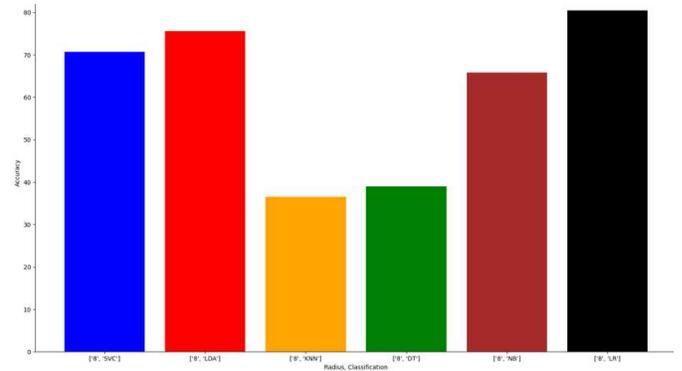

Fig.5. Training_HM vs.Testing_M

*Experiment 6:* In this experiment, the half-masked dataset created in experiment 5 is used to train our 6 ML models. Then, we tested each of the models with 41 masked images, 1 for each one of 41 individuals. Figure 5 shows the accuracy levels of 6 ML models in this experiment.

TABLE 4. MISS RATES FOR MODELS TRAINED WITH UNMASKED FACE IMAGES.

| Miss Rate Table | Dataset: Training_UM | | |
|---|---|---|---|
| ML Algorithm | Group | Misses | Out Of | Percent |
| SVC | Unmasked | 6 | 41 | 15% |
| | Masked | 21 | 41 | 51% |
| LDA | Unmasked | 4 | 41 | 10% |
| | Masked | 16 | 41 | 39% |
| KNN | Unmasked | 9 | 41 | 22% |
| | Masked | 31 | 41 | 76% |
| DT | Unmasked | 20 | 41 | 49% |
| | Masked | 29 | 41 | 71% |
| LR | Unmasked | 1 | 41 | 2% |
| | Masked | 18 | 41 | 44% |
| NB | Unmasked | 7 | 41 | 17% |
| | Masked | 21 | 41 | 51% |

## V. EXPERIMENTAL RESULTS

We used 3 training datasets, 2 testing datasets and 6 Machine Learning Algorithms in our experiments. As described in Table 1, Training_UM dataset contains 369 unmasked images of 41 subjects from ORL database. Training_M also contains 369 images which are all synthesized masked images from the Training_UM dataset using MaskTheFace software. Similarly, Training_HM dataset contains 328 images with 164 masked and remaining 164 unmasked images of 41 subjects with 8 images belonging to each subject. The two testing datasets, Testing_UM and Testing_M contains 41 unmasked and 41 masked images corresponding to 41 subjects. We used Local Binary Pattern (LBP) operator for feature extraction. We set up radius and neighborhood size of LBP to 8 and 24 respectively for each of the 6 machine learning algorithms. The experiments were performed in PyCharm environment. The results are recorded and reported using tables.

TABLE 5. MISS RATES FOR MODELS TRAINED WITH HALF-MASKED FACE IMAGES

| Miss Rate Table | Dataset: Training_HM | | |
|---|---|---|---|
| ML Algorithm | Group | Misses | Out Of | Percent |
| SVC | Unmasked | 8 | 41 | 20% |
|  | Masked | 12 | 41 | 29% |
| LDA | Unmasked | 10 | 41 | 24% |
|  | Masked | 10 | 41 | 24% |
| KNN | Unmasked | 17 | 41 | 41% |
|  | Masked | 26 | 41 | 63% |
| DT | Unmasked | 25 | 41 | 61% |
|  | Masked | 25 | 41 | 61% |
| LR | Unmasked | 8 | 41 | 20% |
|  | Masked | 8 | 41 | 20% |
| NB | Unmasked | 11 | 41 | 27% |
|  | Masked | 14 | 41 | 34% |

As shown in Table 2, the ML models trained with unmasked face images and tested with masked images, LDA is found to be degraded the least with the accuracy of 61%, where as KNN is found to have degraded most with the accuracy of 24%. This result is also illustrated in Figure 3.

The models trained with half-masked and tested with masked images, LR has the highest accuracy of 80% and KNN has the lowest accuracy of 37%, which is also shown in Figure 5. The models trained with masked images and tested with masked images, LR has the highest accuracy of 80% and DT has lowest accuracy of 37% which is also displayed in Figure 4.

TABLE 6. MISS RATES FOR MODELS TRAINED WITH MASKED FACE IMAGES

| Miss Rate Table | Dataset: Training_M | | |
|---|---|---|---|
| ML Algorithm | Group | Misses | Out Of | Percent |
| SVC | Unmasked | 17 | 41 | 41% |
|  | Masked | 11 | 41 | 27% |
| LDA | Unmasked | 15 | 41 | 37% |
|  | Masked | 14 | 41 | 34% |
| KNN | Unmasked | 34 | 41 | 83% |
|  | Masked | 16 | 41 | 39% |
| DT | Unmasked | 30 | 41 | 73% |
|  | Masked | 26 | 41 | 63% |
| LR | Unmasked | 12 | 41 | 29% |
|  | Masked | 8 | 41 | 20% |
| NB | Unmasked | 19 | 41 | 46% |
|  | Masked | 15 | 41 | 37% |

As shown in Table 3, the ML models trained with unmasked face images and tested with masked images, LDA is found to have highest F1score of 76% where as KNN is found to have the lowest of 39%.

TABLE 7. OVERALL AVERAGE MISS RATES FOR ALL DATASETS

| Overall Miss Rates for All Datasets | UM | HM | M | Average |
|---|---|---|---|---|
| Average Unmasked Miss Rate | 19.0% | 32% | 52% | 34% |
| Average Masked Miss Rate | 55.0% | 39% | 37% | 44% |
| Averages | 37.0% | 35.5% | 44.5% | 39.0% |

The models trained with half-masked and tested with masked images, LR has the highest F1 score of 89% and KNN has the lowest of 54%. The models trained with masked images and tested with masked images, LR has the highest F1 score of 89% and DT has lowest score of 54%.

In Table 2, we notice that the highest average performance is 81% when dataset is trained with unmasked images and tested with unmasked images. This is understandable because these ML algorithms are tuned to work with unmasked face images. We observe that the lowest average performance is 45% when system is trained with unmasked faces and tested with masked faces. This shows that models trained with unmasked faces are not suitable for testing with masked faces.

In Table 2, we see that the average accuracy of ML models decrease for testing masked face images when trained with either unmasked or half-masked images. The accuracy is found to be increased for testing masked face images when the ML model is trained with masked faces. In tables 2, 4, 5 and 6, we see that as accuracy decreases the miss rate increases, and vice versa.

As shown in Table 2, LR is found to outperform other models in identifying unmasked facial images for all 3 types of trainining datasets. LDA outperforms other models for identifying masked face images when trained with unmasked face images, LR outperforms other models for identifying masked images when trained with masked or half-masked images.

If a system needs recognizing both masked and unmasked images, then the suggested configuration, as shown in Table 2, is to train with half-masked face images and to use LR as the ML classification model. In this scenario, LR performs with an average of 80% accuracy. If a system needs recognizing only unmasked face images, then the best configuration would be training with unmasked face images and using LR ML model, as Table 2 shows 98% accuracy in this scenario. If there is a need to trecognize only masked face images, then the best performance is obtained by training with masked faces, and using LR model for classification, as LR has an accuracy of 80% in this scenario as shown in Table 2.

In table 7, we notice that while testing masked face images, models trained with masked face images have the lowest average miss rate of 37%, while the models trained with unmasked face images have the highest average miss rates of 55%.

VI. CONCLUSIONS

We observe that LR is the best performing ML model for a system to recognize both masked and unmasked face images by training the model with half-masked image dataset.

We see that LR is the best performing ML model with an accuracy of 98% for a system to recognize only unmasked face images by training the model with unmasked image dataset. In this scenario, DT has an accuracy of 51% making it worst for recognizing unmasked face images.

We notice a trend of increase in accuracy for identifying masked face images for the ML models trained with more masked face images while a trend of decrease is observed the same time for identifying unmasked images.

We will study the difference of real and synthesized masked images in face recognition with deep learning algorithms in the future. We consider working with multiple facial image databases and including databases with large number of subjects as potential future work.

VII. ACKNOWLEDGEMENT

This research is funded by NSF Award #1900087. Any opinions, findings, and conclusions or recommendations expressed in this material are those of the author(s) and do not necessarily reflect the views of NSF.